\documentclass{article}

\usepackage[preprint]{nips_2018}

\usepackage{graphicx}
\usepackage[ruled,vlined]{algorithm2e}
\usepackage{amsmath}
\newcommand{\rpm}{\raisebox{.2ex}{$\scriptstyle\pm$}} 
\usepackage[utf8]{inputenc} 
\usepackage[T1]{fontenc}    
\usepackage{hyperref}       
\usepackage{url}      

\usepackage{natbib}
\setcitestyle{numbers,open={},close={}}

\usepackage{booktabs}       
\usepackage{amsfonts}       
\usepackage{nicefrac}       
\usepackage{microtype}
\usepackage{wrapfig}
\usepackage{multirow}
\usepackage{multicol}

\title{Bayesian Models of Functional Connectomics and Behavior}

\author{%
  Niharika Shimona D'Souza \\
  Department of Electrical and Computer Engineering\\
  Johns Hopkins University\\
  \texttt{ndsouza4@jhu.edu} \\
}

\begin{document}

\maketitle

\begin{abstract}
The problem of jointly analysing functional connectomics and behavioral data is extremely challenging owing to the complex interactions between the two domains. In addition, clinical rs-fMRI studies often have to contend with limited samples, especially in the case of rare disorders. This data-starved regimen can severely restrict the reliability of classical machine learning or deep learning designed to predict behavior from connectivity data. In this work, we approach this problem from the lens of representation learning and bayesian modeling. To model the distributional characteristics of the domains, we first examine the ability of approaches such as Bayesian Linear Regression, Stochastic Search Variable Selection after performing a classical covariance decomposition. Finally, we present a fully bayesian formulation for joint representation learning and prediction. We present preliminary results on a subset of a publicly available clinical rs-fMRI study on patients with Autism Spectrum Disorder.
\end{abstract}

\section{Introduction}

Resting state fMRI (rs-fMRI) is a popular paradigm for assessing brain activity and localize critical functions through steady state patterns of co-activation \cite{fox2007spontaneous}. Network-based approaches to rs-fMRI analysis often group voxels in the brain into regions of interest (ROIs) via a standard anatomical or dervied functional atlas~\cite{nandakumar2018defining},\cite{nandakumarmodified}. From here, the synchrony between the regional time courses can be summarized using a similarity matrix, which can be used as input for further analysis. In the context of neuropsychiatric disorders such as Autism, inter-patient variability often manifests as a spectrum of impairments, that clinicians  quantify as ``behavioral score" of clinical severity obtained from an exam. Identifying sub-networks in the brain that are predictive of such severity can help us understand the social and behavioral implications of the disorder for developing effective behavioral therapy.

Building predictive models at the patient level continues to remain as an open challenge due to the high data dimensionality and considerable inter-subject variation and noise in the resting state acquisition. From a frequentist perspective, predictive models often follow a two step procedure. To combat the data-dimensionality, feature selection is first applied to the raw correlation values i.e. obtained by vectorizing the entries in the similarity matrices. Examples approaches include graph theoretic measures (betweenness,node degree), statistical and/or embedding features obtained from representation learning techniques such as PCA, k-PCA or ICA \cite{khosla2019machine}. Next, regression models such as Random Forests/ Support Vector Regression are applied to the derived features to predict the clinical measures. These strategies have shown success at modeling the group-averaged functional connectivity across the cohort but often fail to accurately capture individual variability. Consequently, the generalization power of these techniques is limited \cite{d2021blending},\cite{d2020joint} \cite{d2019coupled}

In an attempt to address these limitations, recent focus has shifted towards mechanistic network models that are capable of modeling hierarchy onto existing connectivity notions. For example, community detection techniques  are population-level models that are designed to identify interconnected subgraphs within a larger network. These techiniques have refined our understanding of the  organization of complex systems such as brain networks \cite{bardella2016hierarchical},\cite{andersen2014non}. Extensions to Bayesian community detection algorithms \cite{venkataraman2016bayesian},\cite{venkataraman2013connectivity}, \cite{venkataraman2013connectivity}, \cite{venkataraman2015unbiased} have provided valuable insights in characterizing the social and communicative deficits in neurodevelopmental disorders such as schizophrenia and Autism. Unfortunately all of the above focus on group characterizations, and even studies that consider patient variability \cite{heinsfeld2018identification} or hierarchy in \cite{dai2017predicting} have little generalization power on new subjects.

The recent success of network decomposition models \cite{batmanghelich2012generative} in this space largely based on their ability to simultaneously model the patient and group level information. For example, the work of \cite{eavani2015identifying} introduces a common principal components formulation, where multiple rank one matrix outer products capturing the underlying `generative' basis are combined using patient specific coefficients. The sparse basis networks identify meaningful co-activation patterns common to all the patients, and the coefficients model the patient variability. Similar to the joint network optimization model in \cite{d2020joint,d2018generative}, this project explores the `discriminative' nature of these coefficients. Specifically, we estimate clinical severity of every subject first by constructing bayesian regression models which map the subject-coefficients to the behavioral domain once the decomposition is estimated, and then in an end-to-end bayesian model. Through our experiments, we demonstrate the benefit of this joint bayesian formulation in terms of capturing the variability in the cohort, as well as for uncertainty quantification of the estimates.

We have organised this letter as follows \footnote{This work was performed as a final project for graduate level Bayesian Statistics course offered by the Applied Mathematics and Statistics Department at Johns Hopkins University}. We first briefly describe the ASD dataset which we validate on.  Next, our methods section briefly introduces the dictionary decomposition to 
jointly model group-averaged and patient-specific representations, along with the corresponding inference algorithm. From here, we construct two bayesian regression algorithms, the first of the vanilla variety and the second of the variable selection (SVSS) flavour to predict clinical severity from the subject specific coefficients and the estimation algorithms. We compare and this performance to classical penalized linear regression. Finally, we propose a joint heirarchical bayesian model that simultaneously infers the dictionary representation and regression model parameters given the correlation matrices and scores in an end-to-end fashion.

\subsection{Dataset}

\paragraph{\textbf{rs-fMRI Dataset and Preprocessing.}} We evaluate our method on a cohort of $52$ children with high-functioning ASD released as a part of ABIDE \cite{heinsfeld2018identification} from the KKI site. Rs-fMRI preprocessing was performed according to the prevalidated pipeline in \cite{venkataraman2016bayesian} We use the Automatic Anatomical Labeling (AAL) atlas to define $P=116$ cortical, subcortical and cerebellar regions. 
\paragraph{\textbf{Clinical Scores.}} We consider two measures of clinical severity: Autism Diagnostic Observation Schedule (ADOS) total raw score \cite{payakachat2012autism}, which captures the social  and communicative interaction deficits of the patient along with repetitive behaviors (dynamic range: $0$-$30$), and the Social Responsiveness Scale (SRS) total raw score \cite{payakachat2012autism} which characterizes social responsiveness (dynamic range: $70$-$200$).

\section{Methods}

\subsection{Dictionary Learning on rs-fMRI correlation matrices}
\begin{wrapfigure}[12]{R}{0.35\textwidth}
\small
\begin{center}
 \centerline{\fbox{\includegraphics[scale =0.42]{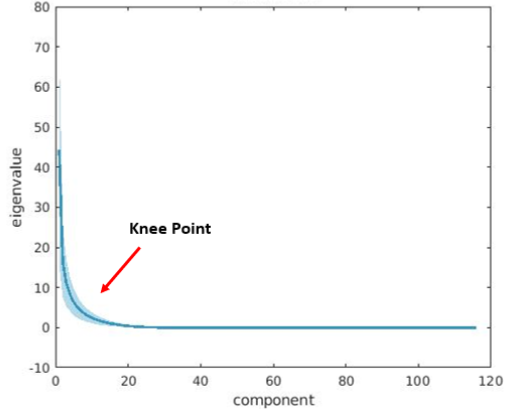}}}
 \medskip 
{\caption{\footnotesize{Scree Plot for $\mathbf{\Gamma}_{n}$}}\label{Scree}}
\end{center}
\end{wrapfigure}
We define $\mathbf{\Gamma}_{n} \in \mathcal{R}^{P \times P}$ as the correlation matrix for patient $n$, where $P$ is the number of regions given by the parcellation. We model $\mathbf{\Gamma}_{n}$ using a group average basis representation and a patient-specific network strength term. The matrix $\mathbf{B} \in \mathcal{R}^{P \times K}$ is a concatenation of  $K$ elemental bases vectors $\mathbf{b}_{k} \in \mathcal{R}^{P \times 1}$, i.e. $\mathbf{B} := \mathbf{b}_{1} \quad \mathbf{b}_{2} \quad ... \quad \mathbf{b}_{K}$, where $K \ll P$. These bases capture steady state patterns of co-activation across regions in the brain. While the bases are common to all patients in the cohort, the combination of these subnetworks is unique to each patient and is captured by the non-negative coefficients $\mathbf{c}_{nk}$. We include a non-negativity constraint $\mathbf{c}_{nk}\geq 0$ on the coefficients to preserve the positive semi-definite structure of the correlation matrices $\{\mathbf{\Gamma}_{n}\}$. The orthonormality constraint on $\mathbf{B}$  helps us learn uncorrelated sub-networks that explain the rs-fMRI data well and implicitly regularize the optimization. Our complete rs-fMRI data representation is:
\begin{equation}
    \mathbf{\Gamma}_{n} \approx \sum_{k}{\mathbf{c}_{nk}\mathbf{b}_{k}\mathbf{b}_{k}^{T}} \ \ \  s.t.  \ \ \ \mathbf{c}_{nk}\geq 0 \ \ \ \ \mathbf{B}^{T}\mathbf{B} =\mathcal{I}_{K}
    \label{eqn1:Eqn1}
\end{equation}
$\mathcal{I}_{K}$ is the $K \times K$ identity matrix. As seen in Eq.~(\ref{eqn1:Eqn1}), we model the heterogeneity in the cohort using a patient specific term in the form of $\mathbf{c}_{n}: = \mathbf{c}_{n1}\quad ... \quad \mathbf{c}_{nK}^{T} \in \mathcal{R}^{K \times 1}$. Taking $\textbf{diag}(\mathbf{c}_{n})$ to be a diagonal matrix with the $K$ patient coefficients on the diagonal and off-diagonal terms set to zero, Eq.~(\ref{eqn1:Eqn1}) can be re-written in matrix form as follows:
\begin{equation}
    \mathbf{\Gamma}_{n} \approx {\mathbf{B}\textbf{diag}({\mathbf{c}}_{n})\mathbf{B}^{T}}
    \ \ \  s.t.  \ \ \ \mathbf{c}_{nk}\geq 0
    \label{eqn:joint}
\end{equation}

Overall, this formulation is similar to common principal components from the statistics and manifold learning literature. Essentially, this strategically reduces the high dimensionality of the data, while providing a patient level description of the correlation matrices. We choose $K=15$ based on the eigenspectrum on $\{\mathbf{\Gamma}_{n}\}$ (See Fig.~\ref{Scree}).
\subsubsection{Optimization}
We use alternating minimization to optimize Eq.~(\ref{eqn:joint}) with respect to $\mathbf{B},\{\mathbf{c}_{n}\}$. Here, we cycle through the updates for the dictionary $\mathbf{B}$, and loadings $\{\mathbf{c}_{n}\}$, to obtain a \textit{joint solution}.
\par We note that there is a closed-form Procrustes solution for quadratic objectives. However, Eq.~(\ref{eqn:joint}) is bi-quadratic in $\mathbf{B}$, so it cannot be directly applied. Therefore, we adopt the strategy in \cite{d2020joint}, by which we introduce the constraints of the form $\mathbf{D}_{n} = \mathbf{B}\mathbf{diag}(\mathbf{c}_{n})$, with corresponding augmented Lagrangian variables $\{\mathbf{\Lambda}_{n}\}$. Thus, our objective from Eq.~(\ref{eqn:joint}) now becomes:

\begin{multline}
\mathcal{J}_{c} = \sum_{n}{\vert\vert{\mathbf{\Gamma}_{n} - \mathbf{D}_{n}\mathbf{B}^{T}}\vert\vert}^{2}_{F}   
+ \sum_{n}{\Big[{\text{Tr}{\left[{(\mathbf{\Lambda}_{n})^{T}({\mathbf{D}_{n}-\mathbf{B}\mathbf{diag}(\mathbf{c}_{n})})}\right]}}+{{\frac{1}{2}}~{\vert\vert{\mathbf{D}_{n}-\mathbf{B}\mathbf{diag}(\mathbf{c}_{n})}\vert\vert}_{F}^{2}}}\Big]  \label{const}
\end{multline}
along with the constraints $\mathbf{c}_{nk} \geq 0$ and  $\mathbf{B}^{T}\mathbf{B} = \mathcal{I}_{K}$. See Algorithm~\ref{Alg:1}

\begin{algorithm}[b]
\SetAlgoLined
\KwResult{Dictionary $\mathbf{B}$ and patient-specific coefficients $\{\mathbf{c}_{n}\}$}
 Initialize $\mathbf{B}^{0}$ as the top eigenvectors of the mean of $\{\Gamma_{n}\}$\;
 \While{Not converged}{
  \textbf{Step 1:} Compute quadratic programming solution for $\{\mathbf{c}_{n}\}$; \\
  \textbf{Step 2:} Compute Procrustes solution for $\mathbf{B}$; \\
  \textbf{Step 3:} Compute Augmented Lagrangian updates for $\{\mathbf{D}_{n},\mathbf{\Lambda}_{n}\}$;
  }
 \caption{Dictionary Learning on rs-fMRI correlation matrices}
 \label{Alg:1}
\end{algorithm}

\subsection{\textbf{Bayesian Regression Models}}
We combine this representation learning with a bayesian regression models to map to clinical severity. Our first set of models consider two classes of Bayesian Regression frameworks to predict behavior from the coefficients $\{\mathbf{c}_{n}\}$.

\subsubsection{Bayesian Linear Regression (BLR)}
Let $\mathbf{y}_{n}$ be the scalar behavioral severity scores for a patient n. We model each $\mathbf{y}_{n} = \mathbf{\beta}_{0} + \mathbf{\beta}^{T} \mathbf{c}_{n} + \epsilon_{n}$, where $\epsilon_{n} \sim \mathcal{N}(0,\sigma^2)$. In this model, we consider that the samples are drawn iid given $\mathbf{c}_{n}$. Our likelihood function is parametrized by $\mathbf{\beta} \in \mathcal{R}^{K\times1} $ and takes the form:
\begin{gather}
    \ell(\{\mathbf{y}_{n}\} \vert \{\mathbf{c}_{n}\},\mathbf{\beta},\mathbf{\beta}_{0},\sigma^2) = \prod_{n=1}^{N}{\ell(\mathbf{y}_{n} \vert \mathbf{c}_{n},\mathbf{\beta},\mathbf{\beta}_{0},\sigma^2)} = \prod_{n=1}^{N}{\mathcal{N}(\mathbf{y}_{n};\mathbf{\beta}^{T}\mathbf{c}_{n},\mathbf{\beta}_{0},\sigma^2)}
    \label{blm}
\end{gather}
We impose a conjugate prior on $(\mathbf{\beta},\sigma^2)$ of the normal inverse-gamma form as follows:
\begin{gather*}
    P(\mathbf{\beta};\beta_{0}\vert\sigma^2) = \mathcal{N}(\mathbf{M},\sigma^2 \mathbf{V}) \ \ \text{and} \ \ \sigma^{2} \sim \text{IG}(a,b)  
\end{gather*}
We set $\mathbf{M}=\mathbf{0} \in \mathcal{R}^{(K+1)\times 1}$ and $a=3,b=1$ as mild assumptions on the prior. For our experiments, we apriori assume that the entries in $\mathbf{\beta}$ are uncorrelated, i.e. $\mathbf{V} = \sigma_{\beta}^2 \cdot \mathcal{I}_{K+1}$. 

\par For our experiments, we consider different values of $\sigma^{2}_{\beta}$ to determine the model with the best fit as a grid search. We use a standard Gibbs Sampling algorithm (Implemented using Matlab's econometrics toolbox) to generate pairs of samples from the posterior $\mathbf{\beta},\mathbf{\beta}_{0},\sigma^2\vert \{\mathbf{c}_{n},\mathbf{y}_{n}\}$ as follows:
\begin{itemize}
    \item[1] Initialize $\mathbf{\beta},\mathbf{\beta}_{0},\sigma^2$
    \item[2] Sample $\mathbf{\beta};\mathbf{\beta}_{0}\vert \{\mathbf{c}_{n},\mathbf{y}_{n}\}, \sigma^2 \sim \mathcal{N}(\mu_{n},\mathbf{\Sigma}_{y} )$ where $\mathbf{\mu}_{n} = (\hat{\mathbf{C}}\hat{\mathbf{C}}^{T} + \sigma^2 \mathbf{V})^{-1}(\hat{\mathbf{C}}\hat{\mathbf{C}}^{T} \hat{\mathbf{\beta}} )$ and 
    $\mathbf{\Sigma}_{y} = (\hat{\mathbf{C}}\hat{\mathbf{C}}^{T} + \sigma^2 \mathbf{V})$, $\hat{\mathbf{C}} = \mathbf{1};\mathbf{C}$
    \item[3] Sample $ \sigma^2 \vert \{\mathbf{c}_{n},\mathbf{y}_{n}\}, \mathbf{\beta};\mathbf{\beta}_{0} \sim \text{IG}\Big(a +{n}/{2},b_{0} + (\sum{\mathbf{y}^{2}_{n}} - \mathbf{\mu}_{n}^{T}\mathbf{\Sigma}_{y}\mathbf{\mu}_{n})/{2} \Big) $
    \item[4] After a burn-in, we keep the rest of the samples as generated from the posterior
\end{itemize}

\subsubsection{Bayesian Stochastic Search Variable Selection (SVSS)}
The goal of variable selection over the linear regression model in Eq.~(\ref{blm}) is to only include only those predictors supported by data in the final regression model. However, analysing $2^{(K+1)}$ permutations of models is computationally inefficient. Instead, Stochastic Variable Selection looks at this problem from a Bayesian perspective. Here, if we wish to exclude a coefficient from a model, we assign it a degenerate posterior distribution that approximates a Dirac-Delta. Practically, if a coefficient is to be included, we draw the coefficient from an $\mathbf{\beta}_{k} \sim \mathcal{N}(\mathbf{0},\mathbf{V}_{1k})$, else $\mathbf{\beta}_{k} \sim \mathcal{N}(\mathbf{0},\mathbf{V}_{2k})$, where $\mathbf{V}_{2k}$ is small relative to $\mathbf{V}_{1k}$. We represent inclusion vs exclusion via a binary variable $\gamma_{k}$, $k =\{0, \dots K\}$. Thus the sample space $\gamma_{k}$ has cardinality $2^{(K+1)}$, and the coefficients $\mathbf{\beta}_{0}, \dots \mathbf{\beta}_{K}$ are independent apriori. Our likelihood model takes the form:
\begin{gather*}
    \ell(\{\mathbf{y}_{n}\} \vert \{\mathbf{c}_{n}\},\mathbf{\beta},\mathbf{\beta}_{0},\sigma^2) = \prod_{n=1}^{N}{\mathcal{N}(\mathbf{y}_{n};\mathbf{\beta}^{T}\mathbf{c}_{n},\mathbf{\beta}_{0},\sigma^2)} \\
    \text{If} \ \ \mathbf{\gamma}_{k} = 1 , \mathbf{\beta}_{k} \sim \mathcal{N}(0,\sigma^2\mathbf{V}_{1k}) \\
    \text{If} \ \ \mathbf{\gamma}_{k} = 0 , \mathbf{\beta}_{k} \sim \mathcal{N}(0,\sigma^2\mathbf{V}_{2k}) 
\end{gather*}
Given $\mathbf{\beta}_{k}$, $\gamma_{k}$ is conditionally independent of the data. Therefore, the full conditional posterior distribution of the probability that the variable $k$ is included in the model.
\begin{gather*}
    P(\mathbf{\gamma}_{k}\vert \mathbf{\beta}_{0},\mathbf{\beta},\sigma^2,\mathbf{\gamma}_{\neq k}) \propto \mathbf{g}_{k} \mathcal{N}(\mathbf{\beta}_{k};0,\sigma^{2} \mathbf{V}_{1k}) \\
    \mathbf{\gamma}_{k} \sim \text{Bernoulli}(\mathbf{g}_{k})
\end{gather*}
We impose a conjugate prior structure on $\mathbf{\beta}_{0};\mathbf{\beta},\{\mathbf{\gamma}\},\sigma^2$ of the following form:
\begin{gather*}
    p(\mathbf{\beta};\mathbf{\beta}_{0},\mathbf{\gamma},\sigma^2) = p(\sigma^2) \prod^{K}_{k=0} p(\beta_{k}\vert\gamma_{k})p(\gamma_{k}) \\  
    \mathbf{\beta}_{k} \vert \mathbf{\gamma}_{k} \sim \mathbb{I}(\gamma_{k}=0) \mathcal{N}(0,\sigma^2\mathbf{V}_{2k}) + \mathbb{I}(\gamma_{k}=1) \mathcal{N}(0,\sigma^2\mathbf{V}_{1k}) \\
    \sigma^2 \sim \text{IG}(a,b)
\end{gather*}
Again, we use a diffuse prior with $a=3,b=1$ and $\mathbf{g}_{k} =0.5$. For our experiments, we consider different values of $\{\mathbf{V}_{1k},\mathbf{V}_{2k}\}$ to determine the model with the best fit as a grid search. We use a standard Gibbs Sampling algorithm (Implemented using Matlab's econometrics toolbox) to generate pairs of samples from the posterior. Although a closed-form posterior exists for conjugate mixture priors, since the prior $\{\mathbf{\beta}\}\vert\sigma^2,\{\gamma_{k}\}$ is marginalized by $\gamma$, this implementation uses MCMC to sample from the joint posterior $\mathbf{\beta},\mathbf{\beta}_{0},\{\mathbf{\gamma}_{k}\} \vert \{\mathbf{c}_{n},\mathbf{y}_{n}\},\sigma^2$. For both of these methods, we run the chains for $5000$ samples as burn in and generate $10000$ additional samples to approximate the posterior.
\begin{wrapfigure}[20]{R}{0.48\textwidth}
\small
\begin{center}
 \centerline{{\includegraphics[scale =0.28]{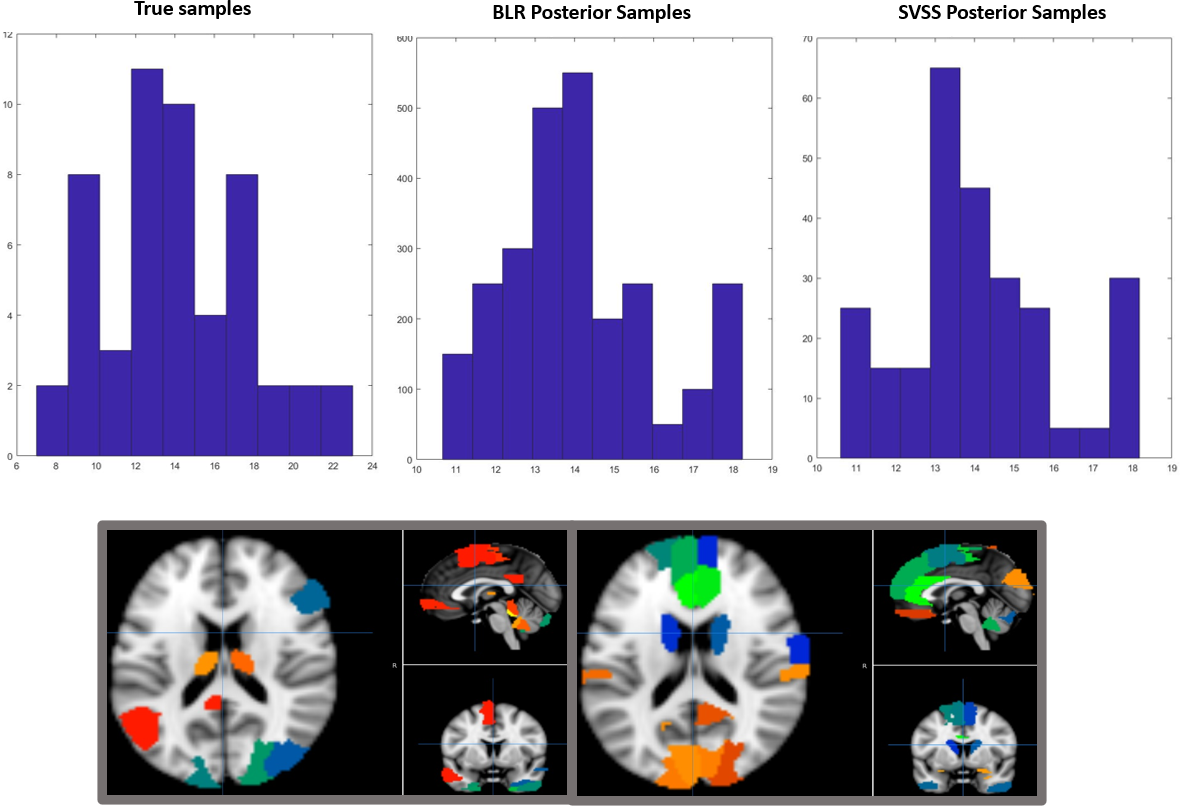}}}
 \medskip 
{\caption{\footnotesize{\textbf{(T)} Histogram for ADOS \textbf{(L)} true samples, \textbf{(M)} posterior from BLR \textbf{(R)} posterior from SVSS 
\textbf{(B)} Subnetworks from SVSS. \textbf{(L)} Visual \& Subcortical \textbf{(R)} Default Mode Network}}\label{SubN}}
\end{center}
\end{wrapfigure}
\begin{figure*}[t!]
   \centering
   \includegraphics[scale=0.49]{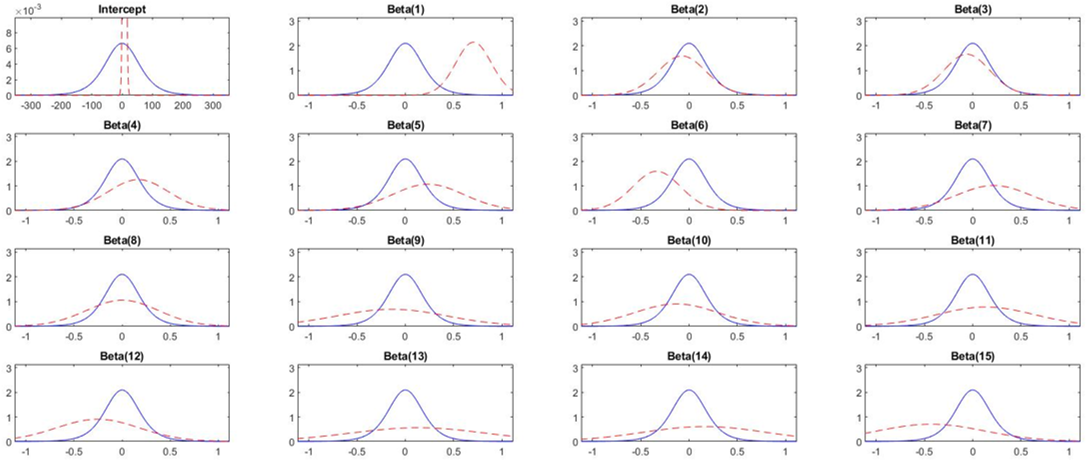}
   \caption{Prior and Posterior Densities for $\mathbf{\beta},\beta_{0}$ by Bayesian Linear Regression on ADOS. The blue line represents the posterior, while the dotted red line is the prior }\label{posterior_densities}
\end{figure*}
\subsubsection{Evaluation and Results}
We evaluate the generalization performance of the model using a five fold cross validation like strategy. In each fold, eighty percent of the examples are used for estimation, and the rest of the twenty percent as forecasting. First, we vary the free parameters $\sigma_{\beta}^2$ for BLR and $\{\mathbf{V}_{1k},\mathbf{V}_{2k}\}$ for SVSS and determine the parameters that fit the estimation points the best. We then evaluate the performance on the held-out forecasting examples. Table~\ref{table:1} compares the performance based on the root mean square error (rMSE) and normalized mutual information (NMI) metric, between the predicted and true samples. SVSS performs better than BLR for both scores. As a baseline, we also report the performance of a classical ridge regression on $\mathbf{c}_{n}$ to predict $\mathbf{y}_{n}$ as well. Using bayesian approaches, we obtain more than a point estimate of each $\mathbf{y}_{n}$ as in frequentist methods. We can also quantify the uncertainty of the estimate around the a posteriori maximum as summary statistics.
\begin{table}[t!]
\footnotesize
\centering
{\caption{\footnotesize{Performance evaluation using \textbf{root Mean Squared Error (rMSE)} \& \textbf{Normalized Mutual Information (MI)}. Lower MAE \& higher MI indicate better performance.}}\label{table:1}}
\begin{tabular}{|c |c | c| c| c|} 
\hline 
  \textbf{Score} &\textbf{Method} &\textbf{rMSE Train} & \textbf{rMSE Test} & \textbf{MI Test}  \\  
\hline 
\hline
  \multirow{3}{4em}{ADOS} &  BLR & 2.76~\rpm~{0.27} & 4.15~\rpm~{0.26} & 0.48 \\
 & SVSS & 2.51~\rpm~{0.45} & 3.80~\rpm~{0.37} & 0.56  \\
 & Ridge Regression & 2.70~\rpm~{2.32} & 3.35~\rpm~{2.11} & 0.41  \\
[0.2ex]  
\hline
 \multirow{3}{4em}{SRS} &  BLR & 34.33~\rpm~{5.44} & 29.23~\rpm~{5.81} & 0.71 \\
 & SVSS & 25.23~\rpm~{5.10}  & 27.99~\rpm~{4.99} & 0.76\\
 & Ridge Regression  & 19.29~\rpm~{11.11}  & 24.44~\rpm~{18.18} & 0.66  \\
 [0.4ex]
 \hline
\end{tabular}
\label{Table:1}
\end{table}
In Fig.~\ref{SubN}, we compare the histograms of the true samples against the samples generated by one run of the BLM and SVSS models for the ADOS score, according to the parameters selected above. The overlap with the true distribution is indicative of how well the models are able to approximate the data generating process. Fig.~\ref{posterior_densities} compares the prior and posterior densities for the coefficients $\mathbf{\beta},\beta_{0}$ learned for BLR. Finally, SVSS allows us to isolate the coefficients $\beta_{k}$ which are consistently selected as predictors. For both ADOS and SRS, $\{\mathbf{c}_{n,7}, \mathbf{c}_{n,8}\}$ are the features that are selected most often (have the highest magnitude of $\beta_{k} > 90 \%$ of the times). We plot the corresponding subnetworks $\mathbf{B}_{k}$ for $k=7,8$ in Fig.~\ref{SubN}. Thus, SVSS offers us interpretability in terms of the features relevant to prediction. Altered connectivity in these networks, both default mode and in visual processing areas has been found to be associated with ASD previously \cite{venkataraman2011joint}.
\begin{wrapfigure}[20]{R}{0.45\textwidth}
\small
\begin{center}
 \centerline{{\includegraphics[scale =0.50]{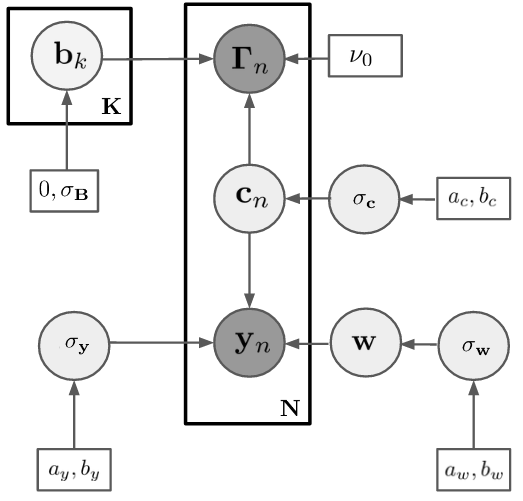}}}
 \medskip 
{\caption{\footnotesize{Hierarchical Bayesian Model for Joint Representation Learning and Prediction}}\label{GM}}
\end{center}
\end{wrapfigure}
\subsection{A Bayesian Model for Joint Representation Learning and Prediction}
In the previous two approaches, the feature extraction is combined with prediction in a pipelined fashion, decoupling the two. However, inherently, the two views of data are complementary to each other. In this section, we propose to use a bayesian model which mimics this representation learning step. At the same time, the patient-specific coefficients relate to prediction via a bayesian linear regression. By combining the dictionary learning directly with prediction, we expect to learn a joint rs-fMRI representation that is more aligned with clinical prediction, similar to the principles in \cite{d2020joint}

\par Recall that the correlation matrices  $\{\mathbf{\Gamma}_{n}\}$ are positive semi-definite. Recall that we use the common principal components decomposition $\mathbf{\Gamma}_{n} \approx \mathbf{B} \mathbf{diag}(\mathbf{c}_{n})\mathbf{B}^{T} $ Additionally, we center $\mathbf{y}_{n}$ to have zero mean. Accordingly, our data likelihood uses an Inverse Wishart distribution ($\mathbf{\Phi}_{W}$) on $\mathbf{\Gamma}_{n}$ centered around $\mathbf{B} \mathbf{diag}(\mathbf{c}_{n})\mathbf{B}^{T}$ with degrees of freedom $\nu_0$. We chose $\nu_0 = P+5$ to center $\{\mathbf{\Gamma}_{n}\}$ loosely around $\mathbf{B}\mathbf{diag}(\mathbf{c}_{n})\mathbf{B}^{T}$. Again, we predict the clinical scores via a linear regression $\mathbf{y}_{n} \approx \mathbf{c}^{T}_{n}\mathbf{w}$ with the linear regression weights $\mathbf{w} \in \mathcal{R}^{K\times 1}$. Let $\theta = (\sigma_{y}^2,\sigma_{c}^2,\sigma_{w}^2)$. Given $\{\mathbf{w},\{\mathbf{c}_{n}\}\}$, we assume that $\mathbf{y}_{n}$ are independent, while given $\{\mathbf{B},\{\mathbf{c}_{n}\}\}$, $\mathbf{\Gamma}_{n}$ are independent. Thus, if $Q =\nu_{0}-P-1$: 
\begin{gather}
    \ell(\{\mathbf{\Gamma}_{n},\mathbf{y}_{n}\} \vert \{\mathbf{c}_{n}\},\mathbf{w},\mathbf{B},\theta) = \prod_{n=1}^{N}{\mathcal{N}(\mathbf{y}_{n};\mathbf{c}_{n}^{T}\mathbf{w},\mathbf{\sigma}_{y}^2)}  \ \mathbf{\Phi}_{W}\Bigg[\mathbf{\Gamma}_{n};\frac{\mathbf{B} \mathbf{diag}(\mathbf{c}_{n})\mathbf{B}^{T}}{Q},\nu_{0}\Bigg]
\end{gather}

We apriori assume that $\mathbf{\Theta} = \{\mathbf{B},\{\mathbf{c}_{n}\},\mathbf{w}\}$ are independent given $\mathbf{\theta}$. Therefore:
\begin{gather}
    P(\mathbf{\Theta},\mathbf{\theta}) = P(\sigma^2_{w}) P(\sigma^2_{c}) P(\sigma^2_{y}) P(\mathbf{w}\vert \sigma^2_{w}) \prod_{k=1}^{K} P(\mathbf{b}_{k})\prod_{n=1}^{N} P(\mathbf{c}_{n}\vert \sigma_{c}^2) 
\end{gather}
To approximate a basis $\mathbf{B}$ that is almost orthogonal [], we use a multivariate normal prior of the form:
\begin{gather}
    P(\mathbf{b}_{k}) = \mathcal{N}(\mathbf{0},\sigma_{B}^2 \mathcal{I}_{P}) \ \ \  \text{s.t.} \ \  \sigma_{B}^2 = \frac{1}{P}
\end{gather}
We use a conjugate multivariate normal-inverse gamma prior on $\{\mathbf{w},\sigma^{2}_{w}\}$ and a half multivariate normal-inverse gamma prior (for non-negativity) on $\{\mathbf{c},\sigma^{2}_{c}\}$ and an inverse gamma prior on $\sigma_{y}^2$:
\begin{gather}
    P(\mathbf{w},\sigma_{w}^{2}) = P(\mathbf{w}\vert\sigma_{w}^{2}) P(\sigma_{w}^{2}) = \mathcal{N}(\mathbf{0},\sigma_{w}^2 \mathcal{I}_{K}) \text{IG}(\sigma_{w}^2;a_{w},b_{w}) \\
    P(\mathbf{C},\sigma_{c}^{2}) = P(\sigma_{c}^{2}) \prod_{n=1}^{N} P(\mathbf{c}_{n}\vert\sigma_{c}^{2})  = \text{IG}(\sigma_{c}^2;a_{c},b_{c}) \prod^{N}_{n=1} \mathcal{N}(\mathbf{0},\sigma_{c}^2 \mathcal{I}_{K}) \\
    P(\sigma^{2}_{y}) = \text{IG}(\sigma_{y}^2;a_{y},b_{y})
\end{gather}
Notice that the complete posterior distributions of $\mathbf{w},\sigma_{w}^2,\sigma_{y}^2,\sigma_{c}^2$ can be derived in closed form owing to the structure of the prior:
\begin{gather}
    \mathbf{w}\vert \mathbf{B},\{\mathbf{c}_{n},\mathbf{\theta}\} \sim \mathcal{N}(\mathbf{w};\mathbf{\mu}_{w},\mathbf{\Sigma}_{w}) \ \ \text{s.t.} \ \ \mathbf{\Sigma}_{w} = \Bigg[ \frac{\mathcal{I}_{K}}{\sigma_{w}^2} + \frac{\sum_{n}{\mathbf{c}_{n}\mathbf{c}^{T}_{n}}}{{\sigma_{y}^2}} \Bigg]^{-1} \ \ \text{and} \ \ \mathbf{\mu}_{n} =  \mathbf{\Sigma}_{w} \frac{\sum_{n}{\mathbf{c}^{2}_{n}}}{\sigma_{y}^2} \label{w} \\
    \sigma^{2}_{w} \vert \mathbf{w},\mathbf{B},\{\mathbf{c}_{n}\},\sigma^{2}_{c},\sigma^{2}_{y} \sim \text{IG}\Bigg(a_{w}+\frac{K}{2},b_{w} + \frac{\sum_{k}\mathbf{w}^{2}_{k}}{2}\Bigg) \label{sig_w} \\
    \sigma^{2}_{c} \vert \mathbf{w},\mathbf{B},\{\mathbf{c}_{n}\},\sigma^{2}_{w},\sigma^{2}_{y} \sim \text{IG}\Bigg(a_{c}+\frac{NK}{2},b_{c} + \frac{\sum_{n}\sum_{k}\mathbf{c}^{2}_{nk}}{2}\Bigg)
    \label{sig_c} \\
    \sigma^{2}_{y} \vert \mathbf{w},\mathbf{B},\{\mathbf{c}_{n}\},\sigma^{2}_{c},\sigma^{2}_{w} \sim \text{IG}\Bigg(a_{y}+\frac{N}{2},b_{y} + \frac{\sum_{n}(\mathbf{c}^{T}_{n}\mathbf{w} -\mathbf{y}_{n})^{2}}{2}\Bigg) \label{sig_y}
\end{gather}
Thus, our inference algorithm performs a Gibbs-MH sampling based on the full conditionals for these variables and random-walk like proposal distributions to sample $\mathbf{B},\{\mathbf{c}_{n}\}$. Our inference algorithm is summarised below (Algorithm~\ref{Alg:2}).  
\begin{algorithm}[b!]
\SetAlgoLined
\KwResult{Posterior samples for $\{\mathbf{B},\{\mathbf{c}_{n}\},\mathbf{w},\mathbf{\theta}\}$}
 Initialize $\mathbf{B}^{0},\{\mathbf{c}^{0}_{n}\},\mathbf{w}^{0},\theta^{0}$ , $a_{c} = a_{w} = a_{y} =3, b_{c} = b_{w} = b_{y} =1$\\
 \While{Not converged}{
  \textbf{Step 1:} Sample $\mathbf{B}^{t} \sim  q(\cdot \vert \mathbf{B}^{t-1})$ ; Determine whether to accept-reject samples \\
  \textbf{Step 2:} Sample $\mathbf{C}^{t} \sim  q(\cdot \vert \mathbf{C}^{t-1})$ ; Determine whether to accept-reject samples \\
  \textbf{Step 3:} Sample $\mathbf{w}$ according to Eqn.~(\ref{w});
  \\ 
  \textbf{Step 4:} Sample $\sigma^{2}_{w}$ according to Eqn.~(\ref{sig_w});
  \\
  \textbf{Step 5:} Sample $\sigma^{2}_{c}$ according to Eqn.~(\ref{sig_c});
  \\
  \textbf{Step 6:} Sample $\sigma^{2}_{y}$ according to Eqn~(\ref{sig_y});
 }
 \caption{Gibbs-MH Sampling for the joint model}
 \label{Alg:2}
\end{algorithm}
\begin{wrapfigure}[19]{H}{0.40\textwidth}
\small
\begin{center}
 \centerline{{\includegraphics[scale =0.30]{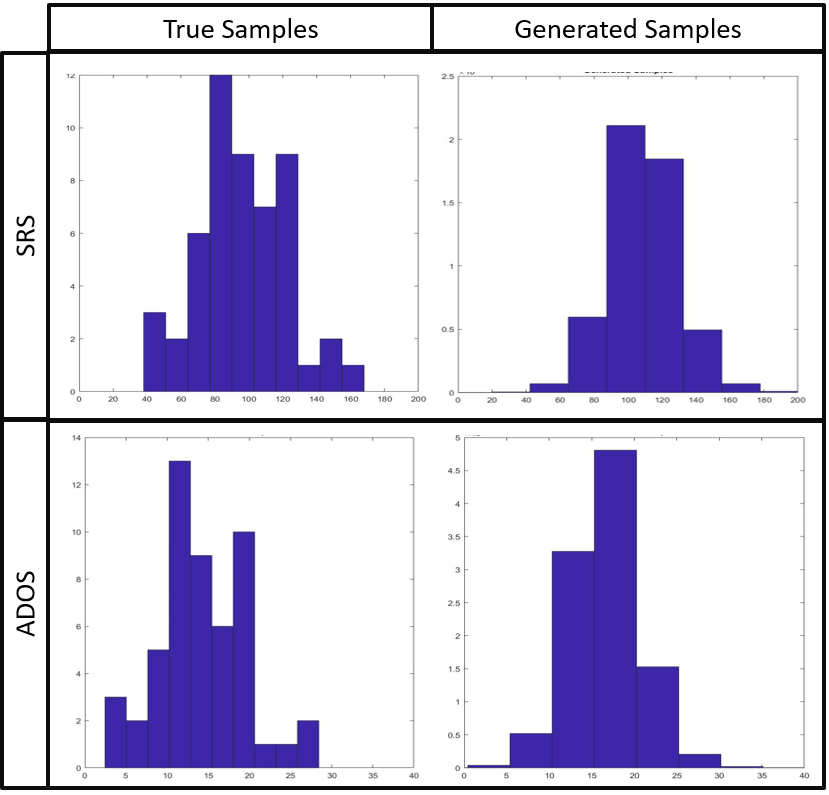}}}
 \medskip 
{\caption{\footnotesize{ Histogram of \textbf{(L)} True samples and \textbf{(R)} Samples from Alg.~\ref{Alg:2}, i.e $\mathbf{c}^{T}_{n}\mathbf{w}$ \textbf{(T)}: SRS \textbf{(B)}: ADOS}}\label{Posterior_Samples_2}}
\end{center}
\end{wrapfigure}

\subsection{Implementation Details and Preliminary Results} We implement Alg.~\ref{Alg:2} in $R$ on an $8$-core machine with an Intel $i7$ processor ($16 $GB RAM). The approximate run time is about 10 hours to generate $10000$ samples. $8000$ of these were treated as burn-in. We experimented with several proposal distributions and found that a normal around the previous sample with a small variance to be the most stable (with acceptance ratio $0.21$ and $0.22$ for the two chains respectively) and concurs with best practices \cite{yildirim2012bayesian}. Also, we fold samples $\mathbf{C}^{t}$ to maintain non-negativity. 
\par
\begin{figure*}[t!]
   \centering
   \includegraphics[scale=0.48]{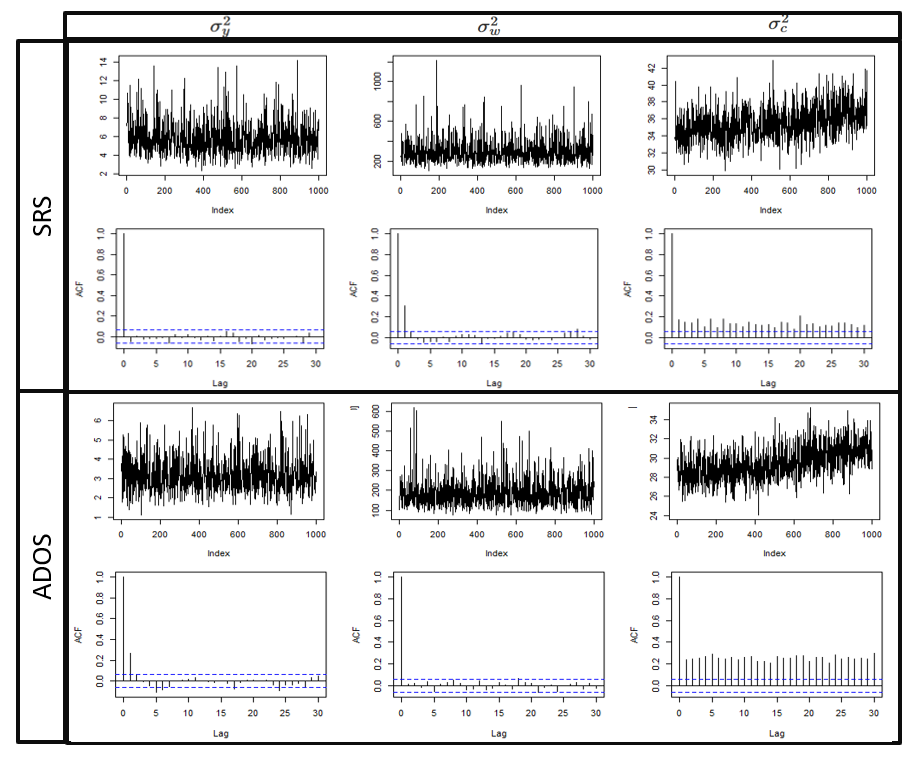}
   \caption{\footnotesize{\textbf{(T)} Trace Plots and \textbf{(B)} Autocorrelation Plots for \textbf{(L)} $\sigma_{w}^2$ \textbf{(M)} $\sigma_{y}^2$
   \textbf{(R)}$~ \sigma_{c}^2$
   where $\mathbf{y}_{n}$ is the SRS score (Top Set) ADOS score (Bottom Set)}
   }\label{ACF}
\end{figure*}

\begin{figure*}[t!]
   \centering
   \includegraphics[scale=0.43]{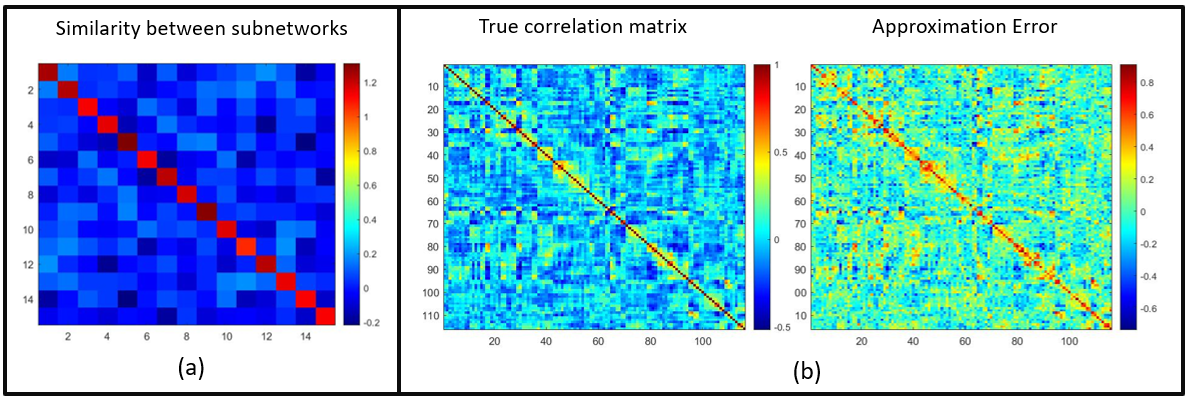}
   \caption{\footnotesize{\textbf{(a)} Inner Product between columns of a representative sample of $\mathbf{B}$ \textbf{(b)} Comparison between a true correlation matrix sample (left), and its mean approximation error (right)}
   }\label{BA}
\end{figure*}

 We first examine the convergence of the chains for $\sigma^2_{w},\sigma^2_{c},\sigma^2_{y}$ via the trace plots and autocorrelation in Fig.~\ref{ACF}. Note that examining the convergence of the other latent variables is a less straightforward exercise. We observed that $\sigma^2_{c}$ has the slowest mixing of these with high autocorrelations between samples, even after running the chains for very long. Additionally, we compare the posterior samples generated by our model i.e. $\hat{\mathbf{y}}_{n} = \mathbf{c}^{T}_{n}\mathbf{w}$ against the distribution of the true scores $\mathbf{y}_{n}$ for both scores. The overlap gives us a sense of how well the generating process is approximated. We obtain an rMSE of $3.78~\rpm~{2.51}$ for ADOS and $19.51~\rpm~{7.51}$
for SRS when using all the samples, which is higher than those obtained in Table~\ref{table:1}. Additionally, as a sanity check, we plot the inner product measure the columns of $\mathbf{B}$ for a representative sample (Fig.~\ref{BA} (a)). Indeed, we see that our chains provide uncorrelated and nearly orthogonal bases. Finally, we plot side by side a correlation matrix sample and the corresponding mean approximation error over samples from the chain (Fig.~\ref{BA} (b)). We notice that while a large number of regions have relatively small approximation errors, the model has trouble determining the prominent patterns along the band diagonal. One of the reasons may be the scale of  $\mathbf{B}$ and $\mathbf{C}$, which is difficult to simultaneously control in a random walk as can be seen with the trace plot for $\sigma^2_{c}$.
\section{Discussion}
In this letter, we first examined the efficacy of  bayesian regression models coupled with dictionary learning to predict clinical severity from rs-fMRI correlation data. Of these, the stochastic variable selection generalized best and offered us with an approach to recognise features most relevant to clinical outcomes. Next, inspired by previous results, we took a fully bayesian approach to jointly learn an rs-fMRI representation and regression model. From the modeling standpoint, this framework presented several design challenges. For example, models selection in terms of order $K$, prior parameters, eg. $(\nu_{0}, a,b)$, currently chosen ad-hoc, and convergence (diagnostics/designing good proposal distributions) of the sampling procedure. Currently, our sampling procedure is computationally expensive and more work needs to be done to ensure that the proposals scan the latent parameter space better and avoid getting stuck in local models. Finally, another challenge is the inherent non-identifiability of the model (for example, scaling- $\{\alpha \mathbf{B},\frac{1}{\alpha^2} \mathbf{C}\}$, $\{\delta{C},\frac{1}{\delta}\mathbf{w}\}$, rotations of $\mathbf{B}$), which contributes to convergence issues when the prior parameters are incorrectly chosen.

\textbf{Future Directions} An immediate future direction could be the extension of the framework to incorporate multimodal structural and functional connectivity data such as~\cite{d2020deep,d2021matrix,dsouza2021m} for behavioral prediction. Another interesting extension of the framework could be towards dynamic modeling of functional connectivity~\cite{d2021deep,nandakumar2020multi,andersen2014non} as it evolves over the scan. Overall, this  preliminary analysis is a first step at exploring the nascent potential of joint bayesian modeling for brain connectivity and behavior. 
{ \bibliographystyle{rusnat}
\bibliography{MyRefs.bib}}

\end{document}